\documentclass[11pt]{article}

\usepackage[final]{acl}

\usepackage{times}
\usepackage{latexsym}

\usepackage[T1]{fontenc}

\usepackage[utf8]{inputenc}

\usepackage{microtype}

\usepackage{inconsolata}

\usepackage{url}
\usepackage{booktabs}
\usepackage{tabularx}
\usepackage{hyperref}   
\usepackage{array}
\usepackage{graphicx}
\usepackage{multirow}
\usepackage{amsmath}
\usepackage[table]{xcolor}
\usepackage{comment}
\usepackage[most]{tcolorbox}

\definecolor{citecolor}{rgb}{0, 0, 0.5}
\definecolor{carebg}{RGB}{232,245,233}
\hypersetup{colorlinks=true, citecolor=citecolor, linkcolor=citecolor, urlcolor=citecolor}

\title{CARE: Privacy-Compliant Agentic Reasoning with Evidence Discordance}

\author{
  \textbf{Haochen Liu\textsuperscript{1}},
  \textbf{Weien Li\textsuperscript{2}},
  \textbf{Rui Song\textsuperscript{2}},
  \textbf{Zeyu Li\textsuperscript{2}},
  \textbf{Chun Jason Xue\textsuperscript{3}},
  \\
  \textbf{Xiao-Yang Liu\textsuperscript{4}},
  \textbf{Sam Nallaperuma-Herzberg\textsuperscript{5,1}},
  \textbf{Xue Liu\textsuperscript{3,2,6}},
  \textbf{Ye Yuan\textsuperscript{2,3,6}}
\\
  \textsuperscript{1}University of Cambridge,
  \textsuperscript{2}McGill University, \\
  \textsuperscript{3}MBZUAI - Mohamed bin Zayed University of Artificial Intelligence,\\
  \textsuperscript{4}Columbia University,
  \textsuperscript{5}King's College London,
  \textsuperscript{6}Mila - Quebec AI Institute
\\
  \small{
    \textbf{Correspondence:} \href{mailto:hl663@cam.ac.uk}{hl663@cam.ac.uk} or \href{mailto:ye.yuan3@mail.mcgill.ca}{ye.yuan3@mail.mcgill.ca}
  }
}

\begin{document}
\maketitle
\begin{abstract}
Large language model (LLM) systems are increasingly used to support high-stakes decision-making, but they typically perform worse when the available evidence is internally inconsistent. 
Such a scenario exists in real-world healthcare settings, with patient-reported symptoms contradicting medical signs. 
To study this problem, we introduce \textbf{MIMIC-DOS}, a dataset for short-horizon organ dysfunction worsening prediction in the intensive care unit (ICU) setting. 
We derive this dataset from the widely recognized MIMIC-IV, a publicly available electronic health record dataset, and construct it exclusively from cases in which discordance between signs and symptoms exists. 
This setting poses a substantial challenge for existing LLM-based approaches, with single-pass LLMs and agentic pipelines often struggling to reconcile such conflicting signals. 
To address this problem, we propose \textbf{CARE}: a multi-stage privacy-compliant agentic reasoning framework in which a proprietary LLM provides guidance by generating structured categories and transitions without accessing sensitive patient data, while a local LLM uses these categories and transitions to support evidence acquisition and final decision-making. 
Empirically, under controlled retrospective evaluation on MIMIC-DOS, CARE achieves the best overall performance across key metrics among the evaluated LLMs and agentic workflows, showing that it can more robustly handle conflicting clinical evidence while preserving privacy.
Our code is \href{https://github.com/AAAAA-Academia-Attractions/CARE}{here}.
\end{abstract}

\section{Introduction}
\label{sec:intro}
%
%
Large language model (LLM) systems are increasingly being used as decision-support tools in high-stakes domains~\citep{singhal2023large, thirunavukarasu2023large}. 
This trend is partly driven by recent advances in prompting, tool use, and agentic orchestration, which have improved LLM performance on complex tasks~\citep{yao2023react, wang2024survey}. 
However, such capabilities are often evaluated in relatively protected settings, where the available evidence is internally consistent and supports a clear decision.
Real-world decision-making, by contrast, frequently involves incomplete, noisy, and conflicting information that must be reconciled before a reliable judgment can be made~\citep{helou2020uncertainty}. 
As a result, it remains unclear whether current LLM systems can reason robustly when the available evidence is internally discordant.

In this work, we focus on cases where subjective and objective evidence point toward different clinical judgments. 
For example, patient-reported symptoms or bedside narrative assessments may appear reassuring, while structured physiological signs indicate risk of deterioration.
Such cases are especially difficult because the correct decision cannot be obtained by naively aggregating all available signals~\citep{helou2020uncertainty}.
Instead, the model should identify what additional information is needed and revise its judgment as new evidence becomes available~\citep{li2024mediq}.
Existing single-pass LLMs and current medical evaluation paradigms are poorly matched to this setting. They are typically studied in static, single-pass formats~\citep{johri2025evaluation}, yet performance becomes noticeably fragile when information must be elicited and integrated over multiple turns~\citep{hager2024evaluation}.

\begin{figure*}[t]
    \centering
    \includegraphics[width=\linewidth, trim=0.2cm 2cm 0.2cm 2cm, clip]{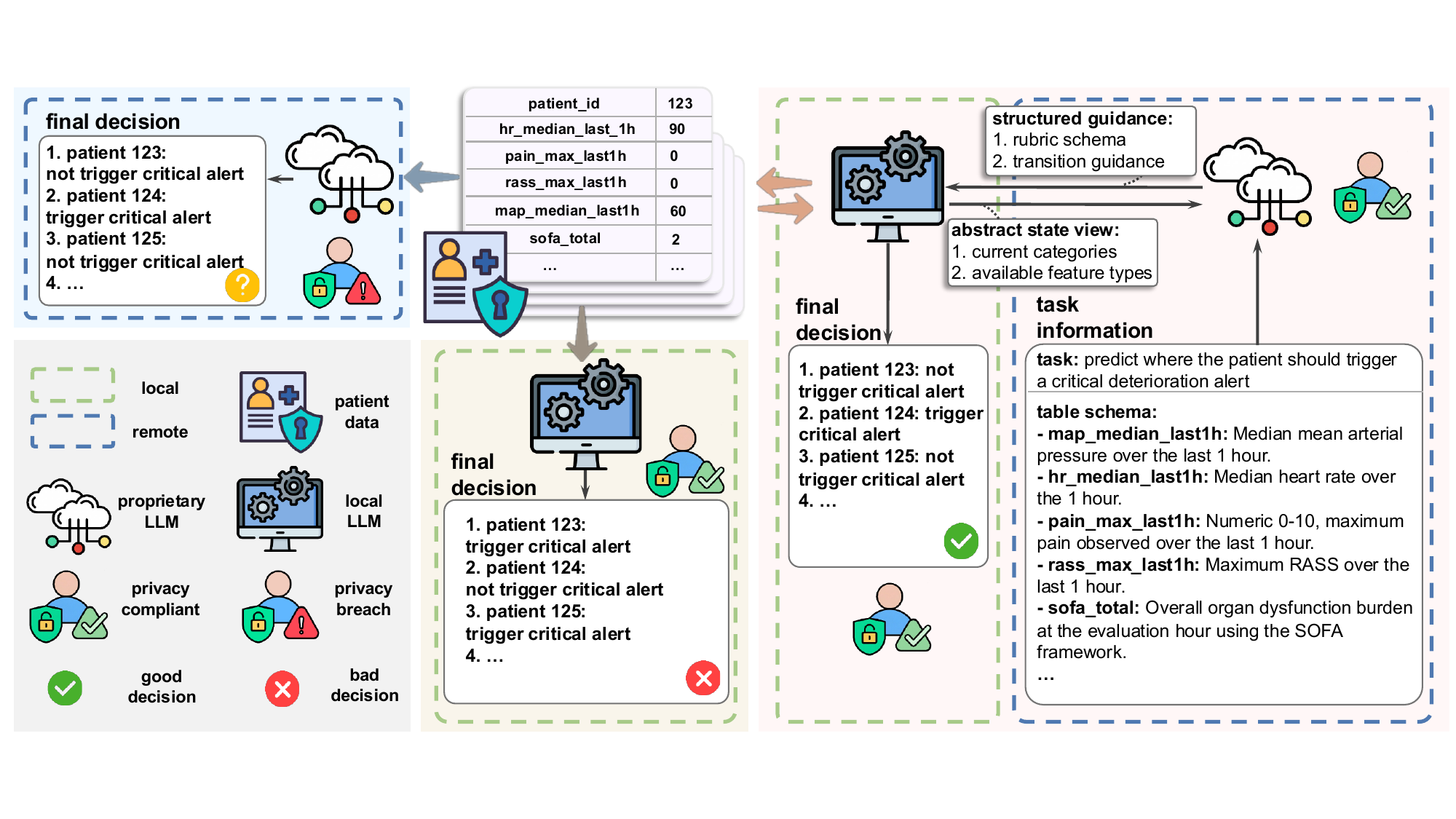}
    \caption{
    Comparison of three decision-making settings: (1) single-pass inference with a proprietary LLM (left, light blue), (2) single-pass inference with a local LLM (middle, light yellow), and (3) our proposed \textbf{CARE} framework (right, light red). 
    The proprietary single-pass setting is non-privacy-compliant because it sends raw patient values to an external model.
    The local single-pass setting keeps patient data local, but its flat one-shot prediction leads to poor performance.
    In contrast, \textbf{CARE} enables the proprietary LLM to provide structured guidance to the local LLM without accessing raw patient values, allowing strong yet privacy-compliant decision making.}
    \label{fig:task}
\end{figure*}

Healthcare settings are further complicated by privacy constraints~\citep{price2019privacy, jonnagaddala2025privacy}. 
In many realistic scenarios, the strongest available model is likely to be a closed-source system that cannot be allowed to access raw patient data, whereas an open-source model can be deployed locally but may still underperform frontier proprietary models on some clinical reasoning tasks~\citep{safavi2025benchmarking, wada2025retrieval}. 
As illustrated in Figure~\ref{fig:task}, this creates a fundamental challenge for our setting: the model best suited for high-level reasoning is not allowed to fully observe the instance, while the model with access to the instance may not reason reliably~\citep{fleming2024medalign}. As a result, end-to-end use of a stronger proprietary model is infeasible, whereas relying on a weaker local model can lead to degenerate decisions~\citep{hager2024evaluation, wada2025retrieval}.

To address this complex challenge while preserving privacy, we propose \textbf{CARE}, a multi-stage privacy-\textbf{C}ompliant \textbf{A}gentic \textbf{RE}asoning framework that separates global guidance from patient-specific reasoning.
Without sending raw patient values to the stronger proprietary model, CARE obtains structured guidance from that model for the reasoning process, including transition-level policies and rubric-level structure derived from feature semantics and task structure.
A local workflow then applies this structure to private patient data for initial state assignment, targeted evidence acquisition, transition evaluation, and final decision-making. 
This decomposition combines the reasoning strength of closed-source models with local patient-specific inference. CARE keeps raw patient values local under the defined privacy constraint and supports disagreement-aware reasoning through explicit state transitions.

To study this setting in a controlled way, we construct \textbf{MIMIC-DOS}, a disagreement-focused benchmark from MIMIC-IV~\citep{PhysioNet-mimiciv-3.1, johnson2023mimic, goldberger2000physiobank} consisting exclusively of cases in which subjective and objective signals point in different directions.  
Since such cases are relatively rare and easily obscured in broader clinical datasets, we isolate them to directly evaluate how LLMs and agentic workflows reason under evidence conflicts.
We further balance this subset to enable stable comparison across methods, while noting that such balance does not reflect real deployment conditions. 
On MIMIC-DOS, several baseline workflows produce highly asymmetric class-wise results, reflected in large differences between TPR and TNR. Such behaviors may appear superficially competitive under aggregated performance metrics, but the highly imbalanced class-wise performance limits their practical utility for clinical decision support.
In contrast, CARE avoids this failure mode and yields more robust performance under evidence discordance.

Our contributions are threefold:
\begin{itemize}
    \item First, we identify sign-symptom discordance as a critical yet underexplored challenge for LLM-based clinical decision support and introduce \textbf{MIMIC-DOS}, a benchmark derived from MIMIC-IV that isolates this mode for controlled evaluation.
    \item Second, we propose \textbf{CARE}, a multi-stage privacy-\textbf{C}ompliant \textbf{A}gentic \textbf{RE}asoning framework that combines structured remote guidance with local patient-specific reasoning under explicit privacy constraints.
    \item Third, we show empirically that evaluated single-pass LLMs and agentic baselines exhibit strong prediction bias toward one class on MIMIC-DOS, while CARE achieves more robust and balanced performance.
\end{itemize}
\section{Related Work}
\label{sec:related}
%
\paragraph{Evidence discordance and the limits of conventional ICU deterioration models}
Predicting patient deterioration in intensive care units (ICUs) has long relied on early warning scores and machine learning models trained on electronic health record datasets such as MIMIC-IV.
Recent studies applied gradient boosting models like XGBoost to predict sepsis mortality, achieving high classification metrics~\citep{li2023developing}.
However, these traditional models often operated as black boxes~\citep{rudin2019stop}.
They flattened complex temporal physiological data into static vectors, offering limited interpretable clinical reasoning to the attending physician, a limitation that drove recent interest in attention-based and inherently interpretable architectures~\citep{choi2016retain}.
More importantly, they struggled to resolve conflicting signals between a patient’s subjective presentation and objective physiological measurements.
This limitation was particularly critical in clinical settings such as occult hypoperfusion (cryptic shock), where patients could appear hemodynamically stable despite elevated risk~\citep{howell2007occult,puskarich2011outcomes}.
In sepsis cohorts, mortality in cryptic shock was reported not to differ significantly from that of overt septic shock~\citep{puskarich2011outcomes}, and elevated lactate has been associated with worse outcomes even in otherwise stable high-risk patients~\citep{meregalli2004occult}.
Together, these findings highlighted a fundamental challenge: clinically important deterioration often manifests as discordance between subjective observations and objective physiological signals, a regime where conventional models are not designed to reason.

\paragraph{LLM and agentic clinical reasoning under conflicting evidence}
Large language models (LLMs) have recently emerged as a promising alternative for clinical decision support.
Models including Med-PaLM demonstrated expert-level proficiency on static medical board examinations~\citep{singhal2023large, nori2023capabilities}.
However, directly applying single LLMs to electronic health records remains challenging due to the dynamic, contradictory, and incomplete nature of clinical data~\citep{jiang2023health}.
Single-pass models were also susceptible to automation bias and misleading textual cues~\citep{goddard2012automation, ji2023survey}, often overweighting reassuring subjective descriptions (e.g., “patient appears calm”) while underweighting subtle but critical physiological abnormalities.
To address these limitations, recent work shifted toward tool-augmented and multi-agent paradigms~\citep{schick2024toolformer, tang2023medagents}.
Frameworks such as ReAct enabled models to iteratively query external tools, grounding reasoning in factual data~\citep{yao2023react}, while multi-agent debate systems attempted to improve robustness through collective reasoning~\citep{du2024improving}.
However, these approaches introduced new challenges in coordinating agents and achieving reliable consensus.
For example, agents could fail to communicate reasoning advantages effectively or converge prematurely to incorrect conclusions~\citep{lin-hooi-2025-enhancing}.
Confidence-aware debate frameworks mitigated this issue by calibrating agent confidence, but they require additional training data, limiting applicability in privacy-constrained and low-resource clinical settings~\citep{lin-hooi-2025-enhancing}.
Concurrent efforts such as FACTS~\citep{yuan2025facts} addressed structured data reliability through offline template generation, separating reasoning logic from data execution.
Building on this insight, we propose an asymmetric hybrid framework that avoids training-dependent calibration by using programmatic evidence checks. 
Specifically, sensitive patient data are processed locally by open-source models, whereas remote proprietary models provide high-level reasoning guidance without direct access to raw data, enabling robust decision-making under evidence discordance while preserving privacy.
\section{Dataset and Task}
\label{sec:dataset_and_task}
%
\paragraph{MIMIC-IV dataset} Our experiments involve MIMIC-IV v3.1~\citep{PhysioNet-mimiciv-3.1, johnson2023mimic, goldberger2000physiobank}, a large de-identified dataset with longitudinal ICU and hospital electronic health records from Beth Israel Deaconess Medical Center between 2008 and 2022. The dataset includes bedside monitoring data, laboratory results, medications, procedures, and clinical documentation, making it suitable for time-indexed ICU prediction tasks.

\paragraph{Construction of MIMIC-DOS}
From MIMIC-IV, we construct \textbf{MIMIC-DOS}, a dataset designed to study ICU states characterized by discordance between bedside subjective presentation and objective physiologic risk. Each sample is defined at the ICU stay–hour level as a pair $(\texttt{stay\_id}, t_{\mathrm{eval}})$, where $t_{\mathrm{eval}}$ denotes a study-defined hourly evaluation time within an ICU stay. Cohort construction is based on three routinely available bedside measures in the one-hour window preceding $t_{\mathrm{eval}}$, including two subjective bedside assessments and one objective hemodynamic measurement. The first is a structured pain self-assessment reported by patients, recorded in MIMIC-IV as an integer-valued bedside pain score ranging from 0 to 10, where 0 indicates no pain~\citep{devlin2018clinical}. The second is the Richmond Agitation-Sedation Scale (RASS), a standard ICU bedside assessment of agitation and sedation ranging from $-5$ (unarousable) to $+4$ (combative), with 0 indicating an alert and calm state~\citep{sessler2002richmond}. The third is mean arterial pressure (MAP), a routinely monitored hemodynamic variable widely used to assess systemic perfusion. A MAP below 65 mmHg is commonly treated as clinically concerning in critical care~\citep{evans2021surviving}.

A sample $(\texttt{stay\_id}, t_{\mathrm{eval}})$ is included only if all of the following conditions are satisfied: (1) the maximum pain score in the preceding hour is 0, indicating no recorded pain; (2) at least one RASS observation is available in the preceding hour, with maximum RASS no greater than 0 and minimum RASS greater than $-3$, excluding agitated and deeply sedated states; and (3) MAP remains below 65 mmHg for a cumulative duration over 5 minutes during the preceding hour. These criteria define ICU stay–hour states in which subjective bedside indicators remain relatively reassuring while objective hemodynamic instability is present.

\paragraph{Task formulation and sample extraction}
MIMIC-DOS is designed for a binary classification task. Each sample $(\texttt{stay\_id}, t_{\mathrm{eval}})$ is associated with a feature set anchored at $t_{\mathrm{eval}}$, consisting of 2 subjective features and 20 objective features. Detailed explanations are in Appendix~\ref{appendix:feature_exposure}. The model predicts whether the patient will experience worsening organ dysfunction within the subsequent 12 hours. Organ dysfunction is operationalized using the MIMIC-IV hourly Sequential Organ Failure Assessment (SOFA) score representation, where each hourly SOFA value is a rolling summary over the preceding 24 hours. A sample is labeled positive if the maximum SOFA score observed from $t_{\mathrm{eval}}+1$ to $t_{\mathrm{eval}}+12$ exceeds the SOFA score at $t_{\mathrm{eval}}$ by at least 2 points, and negative otherwise, following the Sepsis-3 convention that a SOFA increase of at least 2 points indicates clinically meaningful organ dysfunction~\citep{10.1001/jama.2016.0287}. After label assignment, cohort filtering, and overlap exclusion, the eligible evaluation pool contains 25{,}090 ICU stay–hour pairs, including 2{,}179 positive and 22{,}911 negative samples. We then construct a locked benchmark by deterministic pair-level sampling and fix the released evaluation set at 1{,}000 samples, balanced to 500 positive and 500 negative cases. This size keeps evaluation computationally feasible, and the class balance is used for clearer analysis rather than for training, since our workflows operate in a zero-shot setting. The final evaluation set contains 1{,}000 samples from 912 ICU stays and 881 unique patients. 
\section{Method}
\label{sec:method}
%
To solve the challenging problem of discordance between subjective and objective evidence under strict privacy constraints, we introduce \textbf{CARE} (privacy-\textbf{C}ompliant \textbf{A}gentic \textbf{RE}asoning). 
CARE decomposes reasoning into structured stages that coordinate remote guidance and local inference while keeping raw patient measurements local.
In this section, we first define the privacy scope in Section \ref{subsec:privacy_scope} and then describe the details of our proposed CARE framework in Section \ref{subsec:care_framework}.

\subsection{Privacy Scope}
\label{subsec:privacy_scope}
CARE operates under a privacy constraint in which raw patient data remain in the secure local environment and are not provided to the proprietary model. The proprietary model receives only abstract information that does not contain raw patient measurements or sensitive personal information, while all direct processing of patient values is performed locally.

\begin{figure*}[t]
    \centering
    \includegraphics[width=0.9\linewidth, trim=0cm 0cm 0cm 0cm, clip]{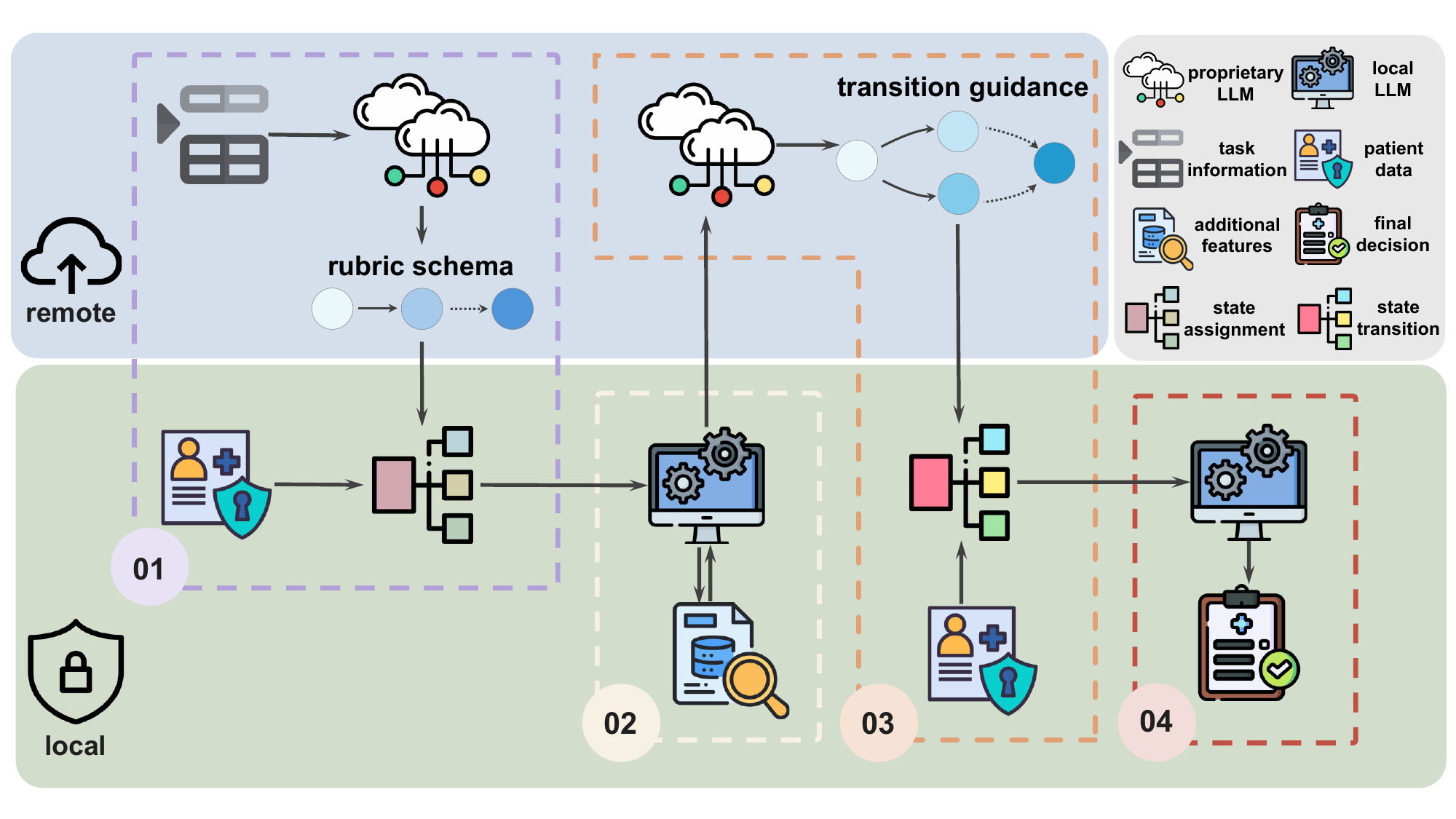}
    \caption{\textbf{Overview of the CARE framework.} 
    In \textbf{Stage 1}, a proprietary LLM constructs a rubric schema over intermediate patient states from task information, and the local side applies this rubric to patient data to obtain an initial state assignment. For each task, the rubric generation process only needs to be done once. 
    In \textbf{Stage 2}, the local LLM performs evidence checks by determining whether additional features are needed given the current state and observed values. 
    In \textbf{Stage 3}, the proprietary LLM generates transition guidance from an abstract view of the current state and available feature types; the local side updates the state through recomputation and constrained merge without exposing raw patient data.
    In \textbf{Stage 4}, the local LLM produces the final task decision from the patient data and the accumulated reasoning trace over states. 
    In our framework, raw patient values remain local throughout the entire pipeline, allowing the framework to preserve privacy.}
    \label{fig:CARE}
\end{figure*}

\subsection{CARE Framework}
\label{subsec:care_framework}
Operationally, CARE consists of four stages, as shown in Figure~\ref{fig:CARE}: 
\textbf{(1) rubric generation and initial state assignment;}
\textbf{(2) category-aware data acquisition;}
\textbf{(3) transition reasoning;}
\textbf{(4) final decision-making.}
The proprietary model generates the rubric from the task description without access to patient-specific data in Stage 1 (rubric generation and initial state assignment) and provides transition guidance in Stage 3 (transition reasoning). Patient-specific state assignment and updating remain local, while the local model handles Stage 2 (category-aware data acquisition) and Stage 4 (final decision-making).
This decomposition allows CARE to combine the abstract reasoning capability of a closed-source model with the defined privacy constraints of local inference and to support disagreement-aware transition-based reasoning. 
Examples of the four stages are in Appendix~\ref{app:stage1}, \ref{app:stage2}, \ref{app:stage3}, and \ref{app:stage4}.

\paragraph{Stage 1: rubric generation and initial state assignment}
In the initial stage, CARE establishes a structured state space for downstream reasoning. 
States are organized as an ordered scale indicating how likely the patient is to be experiencing worsening organ dysfunction.
This rubric schema is generated by an LLM from task-level information, including feature descriptions, feature semantics, and the task objective, without exposing patient-specific values.
The generated rubric defines the intermediate patient states, their semantic interpretations, and the types of evidence typically required to support each state.
In this way, the rubric serves as structured prior knowledge that translates the task objective into a clinically meaningful state space.
The rubric schema is generated once for each task and is reused across all inference instances.
The locally available patient data are then mapped onto this rubric to compute an initial category assignment.
By separating rubric generation from patient-specific instantiation, Stage~1 allows CARE to leverage LLM-derived abstract guidance without exposing raw patient values.

\paragraph{Stage 2: category-aware data acquisition}
Given the initial state assignment, CARE next determines whether the currently available evidence is sufficient for reliable decision-making. 
The local model uses the current category, the patient’s observed values, and the category-specific evidence requirements induced by the rubric to identify which additional features should be retrieved. 
Stage~2 uses state-conditioned acquisition, in which the relevance of additional evidence depends on the patient's current inferred state.
After retrieval, CARE re-evaluates evidence sufficiency and determines whether the available information is adequate for transition reasoning. 

\paragraph{Stage 3: transition reasoning}
Once sufficient evidence is obtained, CARE evaluates how the patient’s state should change based on the newly available information. 
The proprietary model receives the current category, the set of available feature types, and the rubric-level transition structure, but still does not observe any raw patient values. 
Based on this abstract view, it produces structured transition guidance together with candidate state transitions that are plausible to consider. 
The local side then combines this advisory output with the actual patient values through local recomputation and constrained merge, determining whether the patient remains in the same category or transitions to a different one.
This stage is central to CARE's design because it separates abstract transition guidance from local state recomputation, using newly acquired patient evidence to explicitly determine whether the current state should be retained or updated under evidence discordance.

\paragraph{Stage 4: final decision-making}
In the final stage, the local model converts the updated state representation into an initial task-level action. 
This decision is made using the accumulated reasoning trace, including the initial category assignment, the acquired evidence, and the locally updated category and its rationale. 
By grounding the final judgment in the structured outputs of the earlier stages, CARE avoids relying on a single free-form inference pass over heterogeneous evidence. 
A local evidence support check is then applied to verify that an escalation action is supported by the available objective evidence. 
The resulting action is used as the final prediction while patient-specific data remain local throughout.



\section{Experiments}
\label{sec:experiments}
%

\begin{table*}[t]
\centering
\setlength{\tabcolsep}{3pt}
\renewcommand{\arraystretch}{0.95}
\resizebox{\textwidth}{!}{%
\begin{tabular}{
>{\raggedright\arraybackslash}m{2.8cm}
>{\raggedright\arraybackslash}m{2.5cm}
>{\centering\arraybackslash}m{2cm}
c
cc
ccc
c
}
\toprule
\multirow{3}{*}{Workflow} &
\multirow{3}{*}{\shortstack[c]{Local\\LLM(s)}} &
\multirow{3}{*}{\shortstack[c]{Proprietary\\LLM}} &
\multicolumn{1}{c}{Validity} &
\multicolumn{2}{c}{Class Performance} &
\multicolumn{3}{c}{Metrics} &
\multicolumn{1}{c}{Efficiency} \\
\cmidrule(lr){4-4} \cmidrule(lr){5-6} \cmidrule(lr){7-9} \cmidrule(lr){10-10}
& & & \shortstack[c]{Valid\\rate} & \shortstack[c]{TPR\\(Recall)} & \shortstack[c]{TNR\\(Specificity)} & BA & G-mean & MCC & \shortstack[c]{Tokens\\/Sample} \\
\midrule

\multirow{3}{*}{Single-pass}
& GPT-OSS & \textemdash & \textbf{1.0000} & 0.1980 & \textbf{0.8340} & 0.5160 & 0.4064 & 0.0415 & 1155.47 \\
& Qwen    & \textemdash & 0.9980 & 0.3800 & 0.6004 & 0.4902 & 0.4777 & -0.0201 & \textbf{745.42} \\
& LLaDA   & \textemdash & 0.9810 & \textbf{0.9388} & 0.0692 & 0.5040 & 0.2550 & 0.0162 & 834.67 \\
\midrule

Majority voting
& GPT-OSS + Qwen + LLaDA
& \textemdash
& 0.9940 & 0.4289 & 0.5596 & 0.4942 & 0.4899 & -0.0116 & 2739.65 \\
\midrule

RSMAD
& GPT-OSS + Qwen + LLaDA
& \textemdash
& 0.9970 & 0.2751 & 0.7435 & 0.5093 & 0.4523 & 0.0210 & 12912.39 \\
\midrule

ConfMAD
& GPT-OSS + Qwen + LLaDA
& \textemdash
& \textbf{1.0000} & 0.3360 & 0.6620 & 0.4990 & 0.4716 & -0.0021 & 20458.34 \\ 
\midrule

\multirow{3}{*}{CARE (ours)}
& \cellcolor{carebg} GPT-OSS & \cellcolor{carebg} GPT-5 & \cellcolor{carebg} \textbf{1.0000} & \cellcolor{carebg} 0.5220 & \cellcolor{carebg} 0.5700 & \cellcolor{carebg} \textbf{0.5460} & \cellcolor{carebg} \textbf{0.5455} & \cellcolor{carebg} \textbf{0.0921} & \cellcolor{carebg} 7771.52 \\
& \cellcolor{carebg} Qwen    & \cellcolor{carebg} GPT-5 & \cellcolor{carebg} \textbf{1.0000} & \cellcolor{carebg} 0.6520 & \cellcolor{carebg} 0.3560 & \cellcolor{carebg} 0.5040 & \cellcolor{carebg} 0.4818 & \cellcolor{carebg} 0.0084 & \cellcolor{carebg} 7168.53 \\
& \cellcolor{carebg} LLaDA   & \cellcolor{carebg} GPT-5 & \cellcolor{carebg} 0.9990 & \cellcolor{carebg} 0.6232 & \cellcolor{carebg} 0.3660 & \cellcolor{carebg} 0.4946 & \cellcolor{carebg} 0.4776 & \cellcolor{carebg} -0.0111 & \cellcolor{carebg} 7246.58 \\
\bottomrule
\end{tabular}%
}
\caption{Performance comparison across workflows, local LLMs, and proprietary LLMs.}
\label{tab:main_results}
\end{table*}

To structure the empirical evaluation, we study three research questions:
\textbf{RQ1:} How does CARE compare with representative single-pass and multi-agent baseline workflows on MIMIC-DOS?
\textbf{RQ2:} How do different local LLMs affect workflow performance?
\textbf{RQ3:} How do the stages of CARE contribute to its final performance? 

We evaluate CARE on MIMIC-DOS against multiple baseline LLM and agentic workflows. For candidate local LLMs, we consider \textbf{gpt-oss-120b (GPT-OSS)}~\citep{openai2025gptoss120bgptoss20bmodel}, \textbf{Qwen3.5-122B-A10B (Qwen)}~\citep{qwen3.5}, and \textbf{LLaDA2.1-Flash (LLaDA)}~\citep{bie2026llada21speedingtextdiffusion}, spanning two strong autoregressive LLMs and a diffusion-based LLM, allowing us to check how workflows perform across different generation paradigms. 
We use \textbf{GPT-5}~\citep{singh2025openaigpt5card} as the proprietary LLM to support the CARE workflow. 
All experiments are conducted using each LLM's default decoding settings; we do not manually tune temperature or other sampling parameters.
We include four types of baseline workflows in our experiments: 
\textbf{(1)} Single-pass LLM baseline: given the feature block, the model receives all available features of the sample in a single prompt and outputs the final prediction in a single LLM call, without any decomposition, interaction, or iterative refinement. 
\textbf{(2)} Majority voting: three heterogeneous agents based on different local LLMs independently solve the same case without communication~\citep{choidebate}. 
Each agent produces its own final prediction as in the single-pass LLM baseline, and the overall prediction is determined by majority vote over the agents' final outputs. 
\textbf{(3)} Round-synchronous multi-agent debate (RSMAD):
following \cite{du2024improving}, 
each of the three agents first generates an independent answer, as in the single-pass baseline. 
In each of the two subsequent rounds, agents revise their responses using only the responses of the other agents from the previous round, with communication occurring synchronously across rounds. 
The final prediction is determined by majority vote.
\textbf{(4)} Confidence-aware sequential debate (ConfMAD):
following \cite{lin-hooi-2025-enhancing}, 
each of the three agents first produces an independent answer with its reasoning, final prediction, and confidence. 
In the two subsequent rounds, agents respond sequentially in a fixed order, so later agents can observe updates made earlier in the same round. 
The final prediction is determined by selecting the answer with the highest confidence in the last round. Detailed explanations of the two MAD-based baselines are provided in Appendix~\ref{app:MAD_baselines}. 

We evaluate each workflow in terms of \textbf{validity}, \textbf{predictive performance}, and \textbf{efficiency}.
Validity measures whether a workflow returns a valid final output that follows the defined format. Invalid cases are excluded from predictive evaluation.
For predictive performance, we report \textbf{True Positive Rate (TPR)}, \textbf{True Negative Rate (TNR)}, \textbf{Balanced Accuracy (BA)}, \textbf{G-mean}, and \textbf{Matthews Correlation Coefficient (MCC)}.
These metrics are chosen because MIMIC-DOS is a difficult benchmark with subjective--objective discordance, where one-sided prediction collapse can be misleading.
For efficiency, we report \textbf{Tokens/Sample}, the average total token usage per sample.
Metric definitions are provided in Appendix~\ref{app:metric}.

\section{Results and Analysis}
\label{sec:result_and_analysis}

\begin{table*}[t]
\centering
\small
\setlength{\tabcolsep}{3pt}
\renewcommand{\arraystretch}{0.95}

\begin{tabular}{
>{\raggedright\arraybackslash}m{3.2cm}
cc
ccc
c
}
\toprule
\multirow{2}{*}{Workflow} &
\multicolumn{2}{c}{Class Performance} &
\multicolumn{3}{c}{Metrics} &
\multicolumn{1}{c}{Efficiency} \\
\cmidrule(lr){2-3} \cmidrule(lr){4-6} \cmidrule(lr){7-7}
& \shortstack[c]{TPR\\(Recall)} & \shortstack[c]{TNR\\(Specificity)} & BA & G-mean & MCC & \shortstack[c]{Tokens\\/Sample} \\
\midrule

\cellcolor{carebg} Full workflow
& \cellcolor{carebg} 0.5220
& \cellcolor{carebg} 0.5700
& \cellcolor{carebg} \textbf{0.5460}
& \cellcolor{carebg} \textbf{0.5455}
& \cellcolor{carebg} \textbf{0.0921}
& \cellcolor{carebg} 7771.52 \\
\midrule

Backbone only
& 0.3900
& 0.6920
& 0.5410
& 0.5195
& 0.0860
& 4164.42 \\
\midrule

Stage 1 only
& 0.9880
& 0.0020
& 0.4950
& 0.0445
& -0.0600
&  \textemdash \\ 
\midrule

Without Stage 1
& 0.4980
& 0.5660
& 0.5320
& 0.5309
& 0.0641
& 7458.60 \\
\midrule

Without Stage 3
& 0.4160
& 0.6660
& 0.5410
& 0.5264
& 0.0847
& 4202.36 \\
\bottomrule
\end{tabular}%
\caption{Ablation study of CARE under the 4-stage conceptual decomposition. All runs use GPT-OSS as the local model and GPT-5 as the proprietary model. Token counts for rubric schema generation are not included.}
\label{tab:care_ablation}
\end{table*}

\paragraph{Main results}
Table~\ref{tab:main_results} reports the main experimental results. 
Among all compared workflows, \textbf{CARE with GPT-OSS as the local LLM and GPT-5 as the proprietary LLM achieves the best overall performance}, with all outputs being valid and the highest BA (0.5460), G-mean (0.5455), and MCC (0.0921). 
More importantly, it is the only setting in which both the TPR and TNR exceed 0.5, with TPR$=0.5220$ and TNR$=0.5700$.
By contrast, most competing workflows show a pronounced directional bias toward one class.

\paragraph{Workflow-level comparison} 
The baseline workflows exhibit a common failure mode, converging toward model-specific operating points characterized by substantial gaps between TPR and TNR.
This aligns with known vulnerabilities of zero-shot language models to severe label bias and prior probability shifts when forced into immediate classification~\citep{zhao2021calibrate}. 
The issue is most visible in the single-pass setting, where the model receives the full input at once and must directly commit to a final prediction without any intermediate structure for separating evidence acquisition, state formation, and decision making.
By contrast, CARE mitigates this failure mode through its staged agentic framework. 
CARE decomposes the decision process into rubric construction and local state assignment, category-aware evidence acquisition, transition reasoning, and final decision-making. This suggests that CARE's performance reflects the coordinated use of abstract guidance from the proprietary model and local patient-specific reasoning across the workflow.
Majority voting aggregates the agents' binary predictions following the output-level aggregation used in self-consistency protocols~\citep{wang2022self}. In our results, substantial differences between TPR and TNR remain.
The two debate workflow baselines are more expensive but still operate on the same flat input representation without an explicit intermediate state model. 
In our experiments, the two workflows exhibit different trade-offs, with RSMAD favoring specificity and ConfMAD showing relatively higher recall. Both achieve lower BA and MCC than CARE.

CARE differs from the baselines in that it does not make the final decision directly from a flat bundle of mixed evidence. 
This is especially important in our benchmark, which is defined by subjective--objective discordance: reassuring bedside presentation can coexist with genuine physiological risk. 
In this setting, the baseline workflows tend to preserve or merely redistribute model-specific bias, whereas CARE explicitly structures state construction, evidence acquisition, and transition reasoning before the final decision. 
Although CARE consumes more tokens than the single-pass and majority-voting baselines, it achieves the best overall performance on MIMIC-DOS with GPT-OSS as the local LLM and GPT-5 as the proprietary LLM, thereby providing the strongest answer to \textbf{RQ1}.

\paragraph{Comparison by local LLM}
The benefit of CARE is not uniform across local LLMs. 
Relative to the corresponding single-pass baseline, the largest improvement occurs with GPT-OSS, where CARE substantially relaxes the original conservative bias and moves the system toward a much more balanced operating point. 
For Qwen, the gain is smaller. CARE increases positive predictions, but the resulting workflow remains noticeably skewed toward the positive class.
For LLaDA, CARE partially corrects the extreme positive bias of the single-pass baseline, but the final system is still less balanced than the GPT-OSS-based CARE configuration. 
A practical issue with the LLaDA-based workflows is output validity. As shown in Table~\ref{tab:main_results}, they fail to produce valid structured outputs more often than the GPT-OSS- and Qwen-based counterparts.
Overall, these results answer \textbf{RQ2} by showing that local LLM choice materially affects workflow performance, and that CARE is most effective when the local model provides a stable enough prior for the workflow to rebalance its original directional tendency rather than simply amplify it.

\paragraph{Ablation study}
In our 4-stage decomposition, the workflow \textbf{backbone} consists only of Stage~2 (evidence acquisition) and Stage~4 (final decision-making), which together form the minimal executable pipeline. 
Table~\ref{tab:care_ablation} summarizes the ablation results for CARE. 
Relative to the full workflow, the backbone-only variant shows a small but consistent drop in overall performance, indicating that the additional stages provide measurable value beyond this minimal backbone. 

The Stage~1-only (rubric generation and initial state assignment) workflow produces highly one-sided predictions. 
The workflow without Stage~1 causes a further degradation, suggesting that Stage~1, the initial framing step, contributes a useful prior and retrieval direction for subsequent reasoning, evidence acquisition, and state updating.
The workflow without Stage~3 (transition reasoning) leads to a more substantial change in decision behavior. 
It becomes noticeably more conservative, with TPR decreasing from 0.5220 to 0.4160 and TNR increasing from 0.5700 to 0.6660. 
This pattern indicates that Stage~3 integrates the acquired evidence into the final risk judgment and shifts the workflow toward a more balanced class-wise operating point.
Overall, these ablation results answer \textbf{RQ3} by supporting the full staged design of CARE, with Stage~1 providing a modest but useful prior and Stage~3 contributing a more important post-acquisition correction. The best performance is obtained only when all stages are retained.

\paragraph{Supplementary experiments} We also include two supplementary analyses. Appendix~\ref{app:supervised_classifiers} presents supervised learning classifiers as reference points, and Appendix~\ref{app:concordant_evaluation} presents an evaluation under concordant evidence.

\section{Conclusion and Discussion}
\label{sec:conclusion}
We highlight evidence discordance as an important yet underexplored challenge for LLM-based decision-making, and introduce MIMIC-DOS to isolate this case in a controlled yet difficult benchmark. 
We further propose CARE, a privacy-compliant agentic workflow that achieves the strongest overall performance among the evaluated workflows on MIMIC-DOS and, more importantly, is the only evaluated workflow in which both TPR and TNR exceed 50\%, providing benchmark-level evidence of greater robustness than single-pass, voting, and debate-based baselines under controlled retrospective evaluation on MIMIC-DOS.




\section*{Limitations}

Our study has several limitations.
First, MIMIC-DOS focuses on one clinically motivated subtype of subjective-objective evidence discordance and is constructed from a single MIMIC-IV-derived dataset. Its class-balanced design supports controlled comparison but does not reflect natural deployment prevalence. The current results establish CARE's performance in this defined setting. Future work will extend the evaluation to broader discordance types, more natural prevalence distributions, additional clinical tasks, and other healthcare settings.
Second, CARE produces more balanced decisions than the evaluated LLM and agentic workflows, while its absolute predictive performance leaves room for further improvement. Future work will improve overall predictive performance while preserving the balanced decision behavior targeted by CARE.
Third, the current evaluation is retrospective and offline. CARE's design potentially supports online inference. In our implementation, the median end-to-end latency was 53.3 seconds per case, with an interquartile range of 47.0 to 62.4 seconds, corresponding to approximately 0.12\% of the 12-hour prediction horizon. This measured runtime supports completing CARE's inference within the prediction window we studied. Prospective evaluation under dataset shift, changing prevalence, and real-time clinical workflow constraints requires close collaboration with clinical partners and appropriate ethical review and approval. We consider this an important direction for future work.
Finally, the shared rubric is generated once for each task by an LLM and reused across inference instances. It has not yet been designed or systematically validated by clinical experts. Future work will investigate expert-designed and expert-validated rubrics and seek clinician feedback on CARE's validity, feasibility, and clinical utility.

\section*{Ethical Considerations}
\label{sec:ethics_statement}
This study used the MIMIC-IV v3.1 dataset, a de-identified clinical dataset derived from routine clinical care at the Beth Israel Deaconess Medical Center (BIDMC). The collection of patient information and creation of the research resource were reviewed by the Institutional Review Board at BIDMC, which granted a waiver of informed consent and approved the data sharing initiative. Access to the dataset for this work was limited to authorized credentialed users who completed the required human-subjects research training and agreed to the PhysioNet Credentialed Health Data Use Agreement and license terms. The license is the PhysioNet Credentialed Health Data License Version 1.5.0. 
All handling of restricted raw records and all experiments in our study were performed only by credentialed users who completed the required training and signed the data use agreement.
This work is a retrospective secondary analysis of existing de-identified records, does not involve new data collection or patient contact, and reports only aggregate results without disclosing identifiable patient information.

\bibliography{custom}

\newpage
\appendix


\section{Appendix}

\subsection{Features and workflow-specific exposure of MIMIC-DOS}
\label{appendix:feature_exposure}

\begin{table*}
\centering
\small
\setlength{\tabcolsep}{3pt}
\renewcommand{\arraystretch}{1.2}
\resizebox{\textwidth}{!}{%
\begin{tabular}{
>{\raggedright\arraybackslash}p{4.55cm}
>{\raggedright\arraybackslash}p{2.05cm}
>{\raggedright\arraybackslash}p{1.35cm}
>{\raggedright\arraybackslash}p{4.95cm}
}
\toprule
Feature & Clinical domain & Source & Description \\
\midrule

\texttt{pain\_max\_last1h}
& Bedside pain
& Direct
& Maximum bedside pain score. \\

\texttt{rass\_window\_last1h}
& Agitation / sedation
& Derived
& Range-and-count summary of prior-hour RASS observations. \\

\midrule

\texttt{map\_median\_last1h}
& Hemodynamics
& Direct
& Median mean arterial pressure. \\

\texttt{hr\_median\_last1h}
& Hemodynamics
& Direct
& Median heart rate. \\

\texttt{has\_map\_coverage\_last1h}
& Monitoring
& Derived
& Indicator of available MAP coverage. \\

\texttt{map\_covered\_minutes\_last1h}
& Monitoring
& Derived
& Minutes with usable MAP coverage. \\

\texttt{map\_low\_minutes\_last1h\_thr65}
& Hypotension burden
& Derived
& Minutes with MAP below 65 mmHg. \\

\texttt{map\_low\_minutes\_last1h\_thr60}
& Hypotension burden
& Derived
& Minutes with MAP below 60 mmHg. \\

\midrule

\texttt{sofa\_total}
& SOFA
& mimic-code
& Total SOFA score at $t_{\mathrm{eval}}$. \\

\texttt{sofa\_resp}
& SOFA
& mimic-code
& Respiratory SOFA component. \\

\texttt{sofa\_coag}
& SOFA
& mimic-code
& Coagulation SOFA component. \\

\texttt{sofa\_liver}
& SOFA
& mimic-code
& Liver SOFA component. \\

\texttt{sofa\_cardiovascular}
& SOFA
& mimic-code
& Cardiovascular SOFA component. \\

\texttt{sofa\_cns}
& SOFA
& mimic-code
& Central nervous system SOFA component. \\

\texttt{sofa\_renal}
& SOFA
& mimic-code
& Renal SOFA component. \\

\midrule

\texttt{lactate\_latest\_6h}
& Perfusion / metabolism
& Derived
& Most recent lactate value. \\

\texttt{urine\_output\_mlkghr\_6h}
& Renal output
& Derived
& Body-weight-normalized urine output rate. \\

\texttt{norepi\_eq\_dose\_max\_1h}
& Treatment intensity
& Derived
& Maximum norepinephrine-equivalent dose. \\

\texttt{spo2\_latest\_1h}
& Oxygenation
& Derived
& Most recent SpO$_2$ value. \\

\texttt{temperature\_latest\_4h}
& General physiology
& Derived
& Most recent temperature. \\

\texttt{wbc\_latest\_24h}
& Inflammation
& Derived
& Most recent white blood cell count. \\

\texttt{rhythm\_recent\_6h}
& Cardiac rhythm
& Derived
& Most recent rhythm or telemetry summary. \\

\bottomrule
\end{tabular}%
}
\caption{Features used in the workflows. ``Direct'' denotes directly materialized bedside summaries from charted ICU data. ``mimic-code'' denotes SOFA-derived variables inherited from the official MIT-LCP concept definitions. ``Derived'' denotes project-level time-windowed summaries constructed around $t_{\mathrm{eval}}$.}
\label{tab:shared_compact_features}
\end{table*}

The main workflow comparison in this study is conducted on a shared compact feature universe. In this setting, each sample has 22 features in total, including two subjective features and 20 objective features. The two subjective inputs are \texttt{pain\_max\_last1h} and \texttt{rass\_window\_last1h}, which jointly summarize bedside pain and agitation/sedation status in the one-hour window preceding $t_{\mathrm{eval}}$. The 20 objective features cover hemodynamics and hypotension burden, SOFA-based organ dysfunction, perfusion, renal output, treatment intensity, oxygenation, inflammatory context, and rhythm status. Details about all features are listed in Table~\ref{tab:shared_compact_features}.

All baseline workflows operate on the same feature set as a flat evidence bundle. In the single-pass setting, the local model receives the full 22-feature representation and directly produces a final action. Majority voting likewise applies multiple local models to the same full feature bundle and aggregates their outputs. The debate-style baselines also use the same underlying patient representation, differing only in how multiple agents discuss or critique that evidence. 

CARE uses the same 22 features, but does not expose all objective evidence at once. Under the 4-stage conceptual decomposition, the main distinction arises in Stage~2 (category-aware data acquisition), where objective evidence is organized into an initial direct snapshot followed by optional structured retrieval. At the start of Stage~2, the local model has access to a small direct objective snapshot:
\begin{itemize}
    \item \texttt{hr\_median\_last1h}
    \item \texttt{map\_median\_last1h}
    \item \texttt{map\_low\_minutes\_last1h\_thr65}
    \item \texttt{map\_low\_minutes\_last1h\_thr60}
    \item \texttt{has\_map\_coverage\_last1h}
    \item \texttt{sofa\_total}
    \item \texttt{sofa\_cardiovascular}
\end{itemize}

During the remainder of Stage~2, CARE can retrieve additional objective evidence through the structured \texttt{facts\_keys} interface:
\begin{itemize}
    \item \texttt{map\_covered\_minutes\_last1h}
    \item \texttt{lactate\_latest\_6h}
    \item \texttt{urine\_output\_mlkghr\_6h}
    \item \texttt{norepi\_eq\_dose\_max\_1h}
    \item \texttt{sofa\_resp}
    \item \texttt{sofa\_coag}
    \item \texttt{sofa\_liver}
    \item \texttt{sofa\_cns}
    \item \texttt{sofa\_renal}
    \item \texttt{spo2\_latest\_1h}
    \item \texttt{temperature\_latest\_4h}
    \item \texttt{wbc\_latest\_24h}
    \item \texttt{rhythm\_recent\_6h}
\end{itemize}

CARE consumes the same underlying feature universe as the baselines, but organizes objective evidence in a staged manner rather than exposing it all at once. No additional patient features are introduced in Stage~3 (transition reasoning) or Stage~4 (final decision-making). These later stages instead operate on the evidence accumulated earlier in the workflow.

SOFA features play a central role in this compact feature universe. The Sequential Organ Failure Assessment (SOFA) score was originally proposed to describe the severity of organ dysfunction across six organ systems: respiratory, coagulation, liver, cardiovascular, central nervous system, and renal~\citep{vincent1996sofa}. Each component is scored from 0 to 4, with higher values indicating more severe dysfunction, and the total SOFA score is obtained by summing the six components. In our setting, we include both the total SOFA score (\texttt{sofa\_total}) and its component scores (\texttt{sofa\_resp}, \texttt{sofa\_coag}, \texttt{sofa\_liver}, \texttt{sofa\_cardiovascular}, \texttt{sofa\_cns}, and \texttt{sofa\_renal}) so that the workflows can access both overall organ dysfunction severity and its system-level decomposition.

\subsection{MAD Baselines}
\label{app:MAD_baselines}

\paragraph{1. Round-synchronous multi-agent debate}
The \citet{du2024improving} baseline instantiates a 3-agent, 2-round synchronous multi-agent debate. Let the three agents be $A$, $B$, and $C$, and let $R_i^{(t)}$ denote the response of agent $i$ at round $t$. The workflow proceeds as follows.
\begin{enumerate}
\item \textbf{Round 0: independent generation.}
Each agent independently produces an initial response, yielding $R_A^{(0)}$, $R_B^{(0)}$, and $R_C^{(0)}$, conditioned only on the original problem and the patient-parameter block.
\item \textbf{Round 1: synchronous debate.}
For each agent $i$, the system concatenates the complete Round 0 responses of the other two agents, i.e., $\{R_j^{(0)}: j \neq i\}$, and uses them as additional context to generate an updated response $R_i^{(1)}$.
\item \textbf{Round 2: synchronous debate.}
The same mechanism is repeated, except that each agent now only observes the other agents' Round 1 responses, i.e., $\{R_j^{(1)}: j \neq i\}$, to produce $R_i^{(2)}$.
\end{enumerate}

The defining property of this workflow is that updates within each debate round are synchronous. In other words, during Round 1, agent $A$ cannot observe the newly generated Round 1 outputs of $B$ or $C$, and the same holds symmetrically for all agents and for Round 2. This produces a round-synchronous, broadcast-style debate in which all agents update from the preceding round.

In our benchmark, we adopt Round 2 majority vote as the final system output. Under this evaluation protocol, the workflow produces a final action whenever at least two of the three Round 2 agents agree. Cases without a valid majority are treated as vote failures, typically because one agent fails to return a valid output and the remaining two agents disagree.

\paragraph{2. Confidence-aware sequential debate (ConfMAD)}
The \textbf{ConfMAD} baseline implements the plain self-verbalized (SV-Vanilla) variant of a three-agent confidence-aware debate~\citep{lin-hooi-2025-enhancing}. We adopt the SV-Vanilla variant because it preserves the core ConfMAD mechanism of confidence-aware sequential debate, while avoiding additional calibration steps that would require extra held-out data and complicate direct comparison with the other workflows. Unlike in \citet{du2024improving}, each agent in ConfMAD is required to output not only its reasoning and action, but also a confidence score.
Let the output of agent $i$ at round $t$ be
\[
D_i^{(t)} = \bigl(r_i^{(t)}, a_i^{(t)}, c_i^{(t)}\bigr),
\]
where $r_i^{(t)}$ is the reasoning trace, $a_i^{(t)}$ is the final action, and $c_i^{(t)} \in [0,100]$ is the self-reported confidence. The workflow proceeds as follows.
\begin{enumerate}
\item \textbf{Round 0: independent generation.} The three agents independently produce $D_A^{(0)}$, $D_B^{(0)}$, and $D_C^{(0)}$, forming the initial debate history
\[
H_0 = [D_A^{(0)}, D_B^{(0)}, D_C^{(0)}].
\]
\item \textbf{Round 1: sequential debate.} Agents speak in a fixed order $A \rightarrow B \rightarrow C$. First, $A$ reads $H_0$ and generates $D_A^{(1)}$, which is appended to the history.
Then $B$ reads the updated history and generates $D_B^{(1)}$. Finally, $C$ reads the history updated by both preceding agents and generates $D_C^{(1)}$. This yields
\[
H_1 = [H_0, D_A^{(1)}, D_B^{(1)}, D_C^{(1)}].
\]
\item \textbf{Round 2: sequential debate.} The same one-by-one mechanism is repeated. 
Agent $A$ first reads $H_1$ and produces $D_A^{(2)}$, followed by $B$, then $C$, resulting in the final history $H_2$.
\end{enumerate}
The key characteristic is the sequentially updated debate history, with later-speaking agents observing the newly generated outputs, including confidence scores, from earlier agents within the same round. In our implementation, this baseline does not perform confidence calibration and does not introduce a moderator or a judge. Confidence scores are exposed as part of the debate history during sequential interaction. They are also used in the final aggregation, which selects from the valid Round 2 outputs according to the highest confidence score, with deterministic sample-dependent tie-breaking when multiple outputs share the highest confidence score.

\subsection{Metrics}
\label{app:metric}
We evaluate predictive performance using class-balanced metrics derived from the confusion matrix. 
Because our setting requires distinguishing meaningful performance from one-sided prediction collapse, we report balanced accuracy (BA), G-mean, and Matthews correlation coefficient (MCC).

Let $\mathrm{TPR}=\frac{TP}{TP+FN}$ and $\mathrm{TNR}=\frac{TN}{TN+FP}$ denote the true positive rate and true negative rate, respectively. We report the following class-balanced metrics:
$$
\mathrm{BA} = \frac{\mathrm{TPR}+\mathrm{TNR}}{2},
$$
$$
\mathrm{G\text{-}mean} = \sqrt{\mathrm{TPR}\cdot\mathrm{TNR}},
$$

\begin{equation*}
\begin{aligned}
\mathrm{MCC}
&=
\frac{TP\cdot TN - FP\cdot FN}{\sqrt{D}},\\
D
&=
(TP+FP)(TP+FN)\\
&\quad \times (TN+FP)(TN+FN).
\end{aligned}
\end{equation*}
BA measures the average of sensitivity and specificity, giving equal weight to performance on the positive and negative classes. 
G-mean highlights balanced performance across the two classes and strongly penalizes one-sided collapse, since it becomes low when either $\mathrm{TPR}$ or $\mathrm{TNR}$ is low. 
MCC summarizes prediction quality using all four confusion-matrix entries and ranges from $-1$ to $1$, where $1$ indicates perfect prediction, $0$ indicates no better than random association, and $-1$ indicates total disagreement.

\newtcolorbox{promptbox}{
  colback=gray!8,
  colframe=gray!35,
  boxrule=0.4pt,
  arc=2pt,
  left=4pt,
  right=4pt,
  top=4pt,
  bottom=4pt,
  fontupper=\scriptsize,
  enhanced
}

\newtcolorbox{jsonbox}{
  colback=gray!30,
  colframe=gray!35,
  boxrule=0.4pt,
  arc=2pt,
  left=4pt,
  right=4pt,
  top=4pt,
  bottom=4pt,
  fontupper=\scriptsize,
  enhanced
}

\subsection{Illustrative Example of Stage-Wise Data Flow in CARE}
\label{appendix:care_illustrative_example}

To concretely illustrate how CARE operates, we present one representative held-out sample from the official evaluation run. This example is a negative-label case whose final CARE action is \texttt{TREAT\_S}. It is useful because it exhibits the kind of discordance that CARE is designed to handle: a calm bedside presentation together with a mildly concerning objective hemodynamic snapshot. To avoid direct exposure of patient-level MIMIC-IV values, all sample-specific values below are replaced with placeholders. The workflow structure, stage interfaces, requested feature keys, and representative outputs are preserved.

\subsubsection{Stage 1: rubric generation and initial state assignment}\label{app:stage1}
As described in the main workflow definition, Stage~1 establishes the structured state space used for downstream reasoning and maps the currently available local patient data onto that state space to obtain an initial category assignment. In the present experiments, CARE reuses one shared LLM-generated rubric schema for this fixed task. This schema is not patient-specific. It defines the ordered intermediate states used for local initialization and also serves as shared schema-level context for the later transition-reasoning stage. An abbreviated rubric-schema excerpt is shown below. 

{\footnotesize\textit{Representative rubric-schema excerpt.}}
\begin{jsonbox}
\begin{verbatim}
{
  "rubric_schema": [
    {
      "name": "VERY_LIKELY_WORSENING",
      "severity": 5,
      "description": "Clear evidence of active
      deterioration across multiple objective
      domains."
    },
    {
      "name": "LIKELY_WORSENING",
      "severity": 4,
      "description": "Strong concern for worsening,
      but with less complete cross-domain
      confirmation."
    },
    {
      "name": "POTENTIAL_OCCULT_SHOCK",
      "severity": 3,
      "description": "Subjectively calm appearance,
      but objective deviations suggest possible
      occult instability."
    },
    {
      "name": "LIKELY_STABLE",
      "severity": 2,
      "description": "No strong evidence of active
      worsening, though limited abnormalities may
      still be present."
    },
    {
      "name": "VERY_LIKELY_STABLE",
      "severity": 1,
      "description": "Calm bedside state and no
      clear evidence of active deterioration."
    }
  ]
}
\end{verbatim}
\end{jsonbox}

For this sample, the Stage~1 local inputs are represented below using placeholders:

{\footnotesize\textit{Representative local inputs.}}
\begin{jsonbox}
\begin{verbatim}
pain_max_last1h              = <...>
rass_max_last1h              = <...>
rass_min_last1h              = <...>
rass_n_last1h                = <...>
hr_median_last1h             = <...>
map_median_last1h            = <...>
map_low_minutes_last1h_thr65 = <...>
map_low_minutes_last1h_thr60 = <...>
has_map_coverage_last1h      = <...>
sofa_total                   = <...>
sofa_cardiovascular          = <...>
\end{verbatim}
\end{jsonbox}

For each inference instance, the initial state assignment in Stage~1 is computed locally and programmatically. The local side applies the shared rubric schema to the currently available patient data and returns the initial state assignment shown below.

{\footnotesize\textit{Representative JSON output.}}
\begin{jsonbox}
\begin{verbatim}
{
  "matched": true,
  "category": "VERY_LIKELY_STABLE",
  "severity": 1,
  "reason": "Fallback to VERY_LIKELY_STABLE
  (No specific threshold met)."
}
\end{verbatim}
\end{jsonbox}

This output provides the initial category assignment that is passed into Stage~2. In this example, the bedside presentation is calm enough that the sample begins in a low-risk rubric state.

\subsubsection{Stage 2: category-aware data acquisition}\label{app:stage2}
Given the initial category assignment from Stage~1, Stage~2 determines whether the currently available evidence is sufficient for reliable decision-making. Consistent with the Stage-2 design in the main workflow description, the local model conditions its acquisition decision on three elements: the current rubric state, the locally observed patient values, and the evidence requirements implied by that state. 

A representative prompt excerpt is shown below.

{\footnotesize\textit{Prompt excerpt.}}
\begin{promptbox}
\begin{verbatim}
You are an expert ICU AI Triage Agent.

### Locally Available Objective Snapshot
These are already available before any new retrieval.
- MAP median (last 1h):
    {{ map_median_last1h }}
- MAP <65 minutes (last 1h):
    {{ map_low_minutes_last1h_thr65 }}
- MAP <60 minutes (last 1h):
    {{ map_low_minutes_last1h_thr60 }}
- MAP coverage present (last 1h):
    {{ has_map_coverage_last1h }}
- SOFA total: {{ sofa_total }}
- SOFA cardiovascular: {{ sofa_cardiovascular }}

### Current Programmatic State
- Current Category: {{ current_category }}
- Rationale: {{ current_category_reason }}

### Task: Data Acquisition Planning
Review the current state and determine what additional
objective data is needed before decision-making.
\end{verbatim}
\end{promptbox}

For this sample, the local acquisition output is:

{\footnotesize\textit{Representative JSON output.}}
\begin{jsonbox}
\begin{verbatim}
{
  "need_data": true,
  "facts_keys": [
    "map_median_last1h",
    "lactate_latest_6h",
    "urine_output_mlkghr_6h",
    "norepi_eq_dose_max_1h"
  ],
  "reasoning": "Although the programmatic
  rubric currently labels the patient
  VERY_LIKELY_STABLE, the available objective
  snapshot already suggests a possible
  hypotensive burden. Additional cross-domain
  evidence is needed before final
  decision-making."
}
\end{verbatim}
\end{jsonbox}

This output is then checked locally for sufficiency before retrieval:

{\footnotesize\textit{Representative JSON output.}}
\begin{jsonbox}
\begin{verbatim}
{
  "is_sufficient": true,
  "remaining_requested_keys": [],
  "updated_available_keys": [
    "map_median_last1h",
    "lactate_latest_6h",
    "urine_output_mlkghr_6h",
    "norepi_eq_dose_max_1h"
  ]
}
\end{verbatim}
\end{jsonbox}

The requested facts are then materialized from the local feature store:

{\footnotesize\textit{Representative retrieved values.}}
\begin{jsonbox}
\begin{verbatim}
map_median_last1h      = <...>
lactate_latest_6h      = <...>   # or NA
urine_output_mlkghr_6h = <...>   # or NA
norepi_eq_dose_max_1h  = <...>   # or NA
\end{verbatim}
\end{jsonbox}

This trace illustrates Stage~2 as a category-aware acquisition step in which the initial category and locally available evidence guide the selection of relevant objective features across several clinical domains before transition reasoning.

\subsubsection{Stage 3: transition reasoning}\label{app:stage3}
Once sufficient evidence is available, Stage~3 evaluates how the patient state should change in light of the newly available information. In this stage, the proprietary model sees only an abstract view of the case: the current category, the available feature types, and rubric-level transition context. It does not receive raw patient values. Its role is to produce transition guidance and plausible candidate transitions, which the local side then combines with the actual retrieved values through local recomputation and constrained merge.

A representative remote prompt excerpt is shown below.

{\footnotesize\textit{Prompt excerpt.}}
\begin{promptbox}
\begin{verbatim}
You are a Clinical Risk Transition Analyst.

### PRIVACY NOTICE
You are a REMOTE module. You will NOT receive
any actual patient measurements or values.
You will only receive:
- The patient's current risk category
- The types of clinical data that have been
  collected
- Shared rubric-level category definitions

### Current Patient State
- Current Category: {{ current_category }}

### Available Evidence Types
- map_median_last1h
- lactate_latest_6h
- urine_output_mlkghr_6h
- norepi_eq_dose_max_1h

### Category Definitions
...
\end{verbatim}
\end{promptbox}

For this sample, the remote advisory output is:

{\footnotesize\textit{Representative JSON output.}}
\begin{jsonbox}
\begin{verbatim}
{
  "transition_candidates": [
    "VERY_LIKELY_WORSENING",
    "LIKELY_WORSENING",
    "POTENTIAL_OCCULT_SHOCK",
    "LIKELY_STABLE"
  ],
  "transition_guidance": "Prioritize cross-
  domain convergence. Escalate only when
  hemodynamic abnormality is supported by
  corroborating perfusion, support-intensity,
  or organ-failure evidence rather than by a
  single isolated signal.",
  "transition_reasoning": "Starting from
  VERY_LIKELY_STABLE, the most informative
  checks are for early or active shock
  physiology using hemodynamics, support
  intensity, metabolic perfusion, and renal
  perfusion."
}
\end{verbatim}
\end{jsonbox}

Before merging, the local rubric recomputation still returns:

{\footnotesize\textit{Representative JSON output.}}
\begin{jsonbox}
\begin{verbatim}
{
  "matched": true,
  "category": "VERY_LIKELY_STABLE",
  "severity": 1,
  "reason": "Fallback to VERY_LIKELY_STABLE
  (No specific threshold met)."
}
\end{verbatim}
\end{jsonbox}

After local recomputation and constrained merge, the updated state becomes:

{\footnotesize\textit{Representative JSON output.}}
\begin{jsonbox}
\begin{verbatim}
{
  "matched": true,
  "category": "LIKELY_STABLE",
  "severity": 2,
  "reason": "Fallback to VERY_LIKELY_STABLE
  (No specific threshold met).
  [REMOTE_CANDIDATE_MERGE] Local rubric was
  uplifted one level because the retrieved
  pattern supported caution but did not
  provide full cross-domain evidence for
  overt worsening."
}
\end{verbatim}
\end{jsonbox}

\begin{table*}[t]
\centering
\footnotesize
\setlength{\tabcolsep}{4pt}
\renewcommand{\arraystretch}{0.95}
\begin{tabular}{
@{}
>{\raggedright\arraybackslash}m{2.4cm}
ccccc
@{}
}
\toprule
\multirow{2}{*}{Model}
& \multicolumn{2}{c}{Class Performance}
& \multicolumn{3}{c}{Metrics} \\
\cmidrule(lr){2-3}
\cmidrule(lr){4-6}
& \shortstack[c]{TPR\\(Recall)}
& \shortstack[c]{TNR\\(Specificity)}
& BA
& G-mean
& MCC \\
\midrule

XGBoost
& 0.5320
& 0.6600
& 0.5960
& 0.5926
& 0.1936 \\

Random Forest
& 0.5460
& \textbf{0.6680}
& \textbf{0.6070}
& \textbf{0.6039}
& \textbf{0.2156} \\

Decision Tree
& \textbf{0.6680}
& 0.4760
& 0.5720
& 0.5639
& 0.1467 \\
\bottomrule
\end{tabular}

\caption{Performance of supervised learning classifiers on MIMIC-DOS.}
\label{tab:supervised_references}
\end{table*}

This example illustrates the intended role of Stage~3. The remote side supplies transition-level advisory structure, while the actual patient-state update remains local and value-grounded.

\begin{table*}[t]
\centering
\setlength{\tabcolsep}{3pt}
\renewcommand{\arraystretch}{0.95}
\resizebox{\textwidth}{!}{%
\begin{tabular}{
>{\raggedright\arraybackslash}m{2.0cm}
>{\raggedright\arraybackslash}m{2.8cm}
>{\raggedright\arraybackslash}m{1.7cm}
c
cc
ccc
c
}
\toprule
\multirow{2}{*}{Workflow} &
\multirow{2}{*}{\shortstack[c]{Local\\LLM(s)}} &
\multirow{2}{*}{\shortstack[c]{Proprietary\\LLM}} &
\multicolumn{1}{c}{Validity} &
\multicolumn{2}{c}{Class Performance} &
\multicolumn{3}{c}{Metrics} &
\multicolumn{1}{c}{Efficiency} \\
\cmidrule(lr){4-4}
\cmidrule(lr){5-6}
\cmidrule(lr){7-9}
\cmidrule(lr){10-10}
& & &
\shortstack[c]{Valid\\rate} &
\shortstack[c]{TPR\\(Recall)} &
\shortstack[c]{TNR\\(Specificity)} &
BA &
G-mean &
MCC &
\shortstack[c]{Tokens\\/Sample} \\
\midrule

\multirow{3}{*}{Single-pass}
& GPT-OSS & \textemdash
& \textbf{1.0000} & 0.2200 & \textbf{0.8100}
& 0.5150 & 0.4221 & 0.0372 & 1097.70 \\

& Qwen & \textemdash
& \textbf{1.0000} & 0.2200 & 0.7600
& 0.4900 & 0.4089 & -0.0238 & \textbf{747.01} \\

& LLaDA & \textemdash
& 0.9700 & \textbf{0.8454} & 0.1443
& 0.4948 & 0.3493 & -0.0145 & 840.29 \\
\midrule

ConfMAD
& GPT-OSS + Qwen + LLaDA & \textemdash
& \textbf{1.0000} & 0.2200 & 0.7600
& 0.4900 & 0.4089 & -0.0238 & 20553.31 \\
\midrule

\textbf{CARE (ours)}
& \cellcolor{carebg} \textbf{GPT-OSS}
& \cellcolor{carebg} \textbf{GPT-5}
& \cellcolor{carebg} \textbf{1.0000}
& \cellcolor{carebg} 0.6300
& \cellcolor{carebg} 0.5000
& \cellcolor{carebg} \textbf{0.5650}
& \cellcolor{carebg} \textbf{0.5612}
& \cellcolor{carebg} \textbf{0.1311}
& \cellcolor{carebg} 7532.52 \\
\bottomrule
\end{tabular}%
}
\caption{Performance comparison on the concordant-evidence evaluation set.}
\label{tab:concordant_results}
\end{table*}

\subsubsection{Stage 4: final decision-making}\label{app:stage4}
In Stage~4, the local model converts the updated state representation into a task-level action, which is then evaluated by the local evidence support check before being used as the final task-level decision. As described in the main workflow definition, this stage uses the accumulated reasoning trace to produce the final decision from structured evidence gathered across the earlier stages. In this example, the final prompt receives the bedside snapshot, the retrieved facts report, the initial state, and the updated state.

A representative prompt excerpt is shown below.

{\footnotesize\textit{Prompt excerpt.}}
\begin{promptbox}
\begin{verbatim}
You are an expert ICU AI Clinical Triage
Safety Agent.

You requested additional objective clinical
data, and the Laboratory and Monitoring
System returned:

{{ dynamic_facts_report }}

### Heuristic Pre-Assessment (for reference only)
- Initial heuristic estimate: {{ initial_category }}
- Updated heuristic estimate: {{ updated_category }}
- Updated-state rationale: {{ updated_state_reason }}

### Final Triage Decision
You must now make the final clinical triage decision.
\end{verbatim}
\end{promptbox}

For this sample, the final output after the local evidence support check is:

{\footnotesize\textit{Representative JSON output.}}
\begin{jsonbox}
\begin{verbatim}
{
  "differential_diagnosis": "The retrieved
  evidence suggests limited hemodynamic
  concern, but confirmatory perfusion, renal,
  and support signals remain absent or
  unavailable. The overall pattern is not
  sufficient to support escalation to a
  higher-risk action.",
  "final_action": "TREAT_S"
}
\end{verbatim}
\end{jsonbox}

The local evidence support check records:

{\footnotesize\textit{Representative JSON output.}}
\begin{jsonbox}
\begin{verbatim}
{
  "balance_gate": "downgrade_to_treat_s",
  "support_count": 1,
  "support_flags": {
    "hemodynamic": true,
    "perfusion": false,
    "renal": false,
    "pressor": false,
    "organ": false
  }
}
\end{verbatim}
\end{jsonbox}

This final stage shows how CARE converts the updated state and locally accumulated evidence into a conservative but non-escalatory action. For this negative-label sample, the delivered action is \texttt{TREAT\_S}. In the evaluation protocol, the three workflow-level outputs are \texttt{OBSERVE}, \texttt{TREAT\_S}, and \texttt{INVESTIGATE\_O}; for binary deterioration evaluation, \texttt{INVESTIGATE\_O} is treated as the positive prediction, whereas \texttt{OBSERVE} and \texttt{TREAT\_S} are grouped as negative predictions.

\paragraph{Summary of the data flow.}
This example makes the stage-level information flow of CARE explicit. Stage~1 maps local patient data into an initial rubric state using the schema. Stage~2 uses that state to drive category-aware evidence acquisition and local retrieval. Stage~3 exposes only category- and schema-level information to the remote model, receives transition guidance and candidate transitions in structured form, and then updates the state locally through recomputation and constrained merge. Stage~4 converts the updated state and accumulated local evidence into the final task-level action. In this way, the example shows how CARE distributes reasoning across stages while keeping raw patient values local throughout.

\subsection{Supervised Learning Classifiers}
\label{app:supervised_classifiers}

To provide supervised learning references for structured prediction on MIMIC-DOS, we trained XGBoost, Random Forest, and Decision Tree classifiers on a separate balanced 400-sample training set and evaluated them on the 1,000-sample MIMIC-DOS evaluation set. Results are shown in Table~\ref{tab:supervised_references}. These classifiers are evaluated using the same structured objective feature set available to the agentic workflows. No evaluation $(\texttt{stay\_id}, t_{\mathrm{eval}})$ pair appears in the training split. All three classifiers use task-specific labeled training data, whereas CARE and the agentic workflows in the main comparison operate in a zero-shot setting without access to task-specific labeled examples before inference. These results should be interpreted with these different training conditions taken into account.

\subsection{Evaluation under Concordant Evidence}
\label{app:concordant_evaluation}

We also evaluate CARE on a smaller cohort constructed from ICU stay–hour pairs in which subjective bedside indicators and objective hemodynamic measurements both indicate clinical concern. The evaluation set contains 200 samples balanced to 100 positive and 100 negative cases. It uses the same 2 subjective and 20 objective features as MIMIC-DOS and follows the same task definition for predicting worsening organ dysfunction.

As shown in Table~\ref{tab:concordant_results}, CARE achieves the best overall performance among the evaluated workflows on this concordant cohort. It is the only workflow for which both TPR and TNR reach at least 0.5, and it achieves the highest BA, G-mean, and MCC. The single-pass and ConfMAD baselines exhibit pronounced directional bias toward one class, reflected in large differences between TPR and TNR.

\subsection{Potential Risks}

This work is a controlled experimental study conducted on de-identified retrospective ICU data from the MIMIC-IV v3.1 database \citep{PhysioNet-mimiciv-3.1, johnson2023mimic, goldberger2000physiobank}.
The proposed framework and its outputs are intended only for benchmark evaluation and methodological analysis.
They do not constitute medical advice, diagnosis, treatment recommendation, or a real-time clinical decision-support system.
A potential risk is that generated explanations or predicted risk labels may be over-interpreted as clinically actionable recommendations.
In this case, we restrict our claims to offline evaluation, report automated benchmark performance rather than clinical utility, and emphasize that any deployment-facing use would require prospective validation, appropriate governance, and clinician evaluation.

\subsection{LLM Usage}

We used LLM assistants to refine the academic writing and polish the design of the diagrams. The assistance provided by the LLMs did not include method development, experiment execution, or interpretation of the results.

\end{document}